%% file: main.tex
\documentclass[10pt,twocolumn,letterpaper]{article}

\usepackage{tikz}
\usepackage{pgfplots}
\usepackage{stackengine}
\usepackage{graphicx}
\usepackage{subcaption}
\usepackage{booktabs}
\usepackage[accsupp]{axessibility}

\DeclareMathOperator*{\argmax}{arg\,max}
\DeclareMathOperator*{\rank}{rank}

\usetikzlibrary{shapes.geometric}
\usepackage[pagenumbers]{cvpr} 


\definecolor{cvprblue}{rgb}{0.21,0.49,0.74}

\usepackage{enumitem}
\usepackage{amssymb}
\usepackage{mathtools}
\usepackage{dsfont}
\usepackage{amsmath}
\usepackage{appendix}
\usepackage[ruled,vlined]{algorithm2e}
\usepackage[pagebackref,breaklinks,colorlinks,citecolor=cvprblue]{hyperref}

\definecolor{commentcolor}{RGB}{110,154,155}   
\newcommand{\PyComment}[1]{\ttfamily\textcolor{commentcolor}{\# #1}}  
\newcommand{\PyCode}[1]{\ttfamily\textcolor{black}{#1}} 

\def\paperID{13008} 
\def\confName{CVPR}
\def\confYear{2024}

\title{Data-Efficient Multimodal Fusion on a Single GPU}

\author{No\"el Vouitsis\thanks{Authors contributed equally to this work.}  \hspace{1cm} Zhaoyan Liu\footnotemark[1]  \hspace{1cm} Satya Krishna Gorti\footnotemark[1]  \hspace{1cm} Valentin Villecroze \\ Jesse C. Cresswell \hspace{1cm} Guangwei Yu \hspace{1cm} Gabriel Loaiza-Ganem \hspace{1cm} Maksims Volkovs\\ \\
Layer 6 AI\\
{\tt\small \{noel, zhaoyan, satya, valentin.v, jesse, guang, gabriel, maks\}@layer6.ai}
}

\pgfdeclareplotmark{mystarblue}{
	\node[star,star point ratio=2.25,minimum size=6pt,
	inner sep=0pt,draw=black,solid,fill=blue] {};
}
\pgfdeclareplotmark{mystarred}{
	\node[star,star point ratio=2.25,minimum size=6pt,
	inner sep=0pt,draw=black,solid,fill=red] {};
}
\pgfdeclareplotmark{mystargreen}{
	\node[star,star point ratio=2.25,minimum size=6pt,
	inner sep=0pt,draw=black,solid,fill=green] {};
}

\begin{document}
\maketitle
\input{sections/0_abstract}    
\input{sections/1_intro}

\input{sections/2_related}
\input{sections/3_background}

\input{sections/4_motivation}
\input{sections/5_method}
\input{sections/6_experiments}
\input{sections/7_conclusion}
\vspace{-0.85cm}
{
    \small
    \bibliographystyle{ieeenat_fullname}
    \bibliography{main}
}
\input{sections/appendix}


\end{document}

%% file: sections/0_abstract.tex
\begin{abstract}
The goal of multimodal alignment is to learn a single latent space that is shared between multimodal inputs. The most powerful models in this space have been trained using massive datasets of paired inputs and large-scale computational resources, making them prohibitively expensive to train in many practical scenarios. We surmise that existing unimodal encoders pre-trained on large amounts of unimodal data should provide an effective bootstrap to create multimodal models from unimodal ones at much lower costs. We therefore propose FuseMix, a multimodal augmentation scheme that operates on the latent spaces of arbitrary pre-trained unimodal encoders. Using FuseMix for multimodal alignment, we achieve competitive performance -- and in certain cases outperform state-of-the art methods -- in both image-text and audio-text retrieval, with orders of magnitude less compute and data: for example, we outperform CLIP on the Flickr30K text-to-image retrieval task with $\sim\!600\times$ fewer GPU days and $\sim \! 80\times$ fewer image-text pairs. Additionally, we show how our method can be applied to convert pre-trained text-to-image generative models into audio-to-image ones. Code is available at: \url{https://github.com/layer6ai-labs/fusemix}.

\end{abstract}

%% file: sections/1_intro.tex
\section{Introduction}
\label{sec:intro}
Recent advances in multimodal machine learning have unlocked unprecedented capabilities across a wide array of understanding-based \cite{li2021albef, li2022blip} and generation-based \cite{li2023blip2, driess2023palme, liu2023llava, li2023blipdiffusion} applications, some of which have even garnered mainstream attention \cite{ramesh2022hierarchical,rombach2022high, alayrac2022flamingo, yang2023dawn}. 
Of particular interest to us for this work is multimodal alignment \cite{jia2021align, guzhov2022audioclip, girdhar2023imagebind}, which we alternatively refer to as multimodal fusion, wherein the goal is to learn a single latent space that is shared between inputs of various modalities. Recent successes in multimodal fusion have been largely driven by large-scale training regimes requiring many GPUs, and often relying on datasets of billions of multimodal pairs \cite{radford2021clip, yu2022coca, kossen2023three}. This presents a cost that is unacceptable for many practical scenarios where access to compute is limited and where multimodal data is scarce \cite{wang2023too, liu2023tr0n}. It is thus of paramount importance to design efficient frameworks that can democratize research in multimodal fusion.

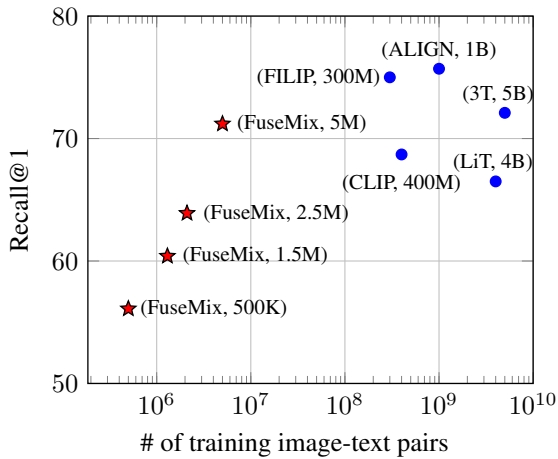
\begin{figure}[t]
\centering
\begin{tikzpicture}
    \begin{axis}[
        width=0.9\linewidth,
	line width=0.5,
        xlabel={\# of training image-text pairs},
        ylabel=Recall@1,
        xmode=log,
        grid=major,
        yticklabel style={
            /pgf/number format/fixed,
            /pgf/number format/precision=1,
        },
        ylabel style={yshift=0cm},
        legend style={
            at={(0.03,0.97)},
            anchor=north west,
            font=\scriptsize,
            draw=none, 
            nodes={scale=0.8, transform shape}
        },
        ymin = 50,
        ymax = 80,
        xmax=10000000000,
        x tick label style={
            /pgf/number format/sci generic={mantissa sep=\times,exponent={10^{##1}},tiny}, 
        },
        ylabel style={yshift=-2ex},
        xlabel style={xshift=-6pt},
    ]

    \addplot[only marks, mark=mystarred,blue] table {performance_data1.dat};

    \addplot[only marks,mark=*, color=blue, solid, mark options={scale=1}] table {performance_data2.dat};
    \node [right, xshift=1] at (axis cs:  500000, 56.1) [font=\footnotesize] {(FuseMix, 500K)};
    \node [right] at (axis cs:  1500000, 60.4) [font=\footnotesize] {(FuseMix, 1.5M)};
    \node [right] at (axis cs:  2500000, 63.9) [font=\footnotesize] {(FuseMix, 2.5M)};
    \node [right, xshift=2] at (axis cs:  5000000, 71.2) [font=\footnotesize] {(FuseMix, 5M)};
    \node [left] at (axis cs:  300000000, 75) [font=\footnotesize] {(FILIP, 300M)};
    \node [below, yshift=-4] at (axis cs:  400000000, 68.7) [font=\footnotesize] {(CLIP, 400M)};
    \node [above] at (axis cs: 1000000000, 75.7) [font=\footnotesize] {(ALIGN, 1B)};
    \node [above, xshift=-1] at (axis cs: 4000000000, 66.5) [font=\footnotesize] {(LiT, 4B)};
    \node [above, xshift=-2.5] at (axis cs: 5000000000, 72.1) [font=\footnotesize] {(3T, 5B)};
    
    \end{axis}
    \end{tikzpicture}
    \vspace{-10pt}
    \caption{Text-to-image retrieval performance as a function of the number of image-text pairs used during training, evaluated on the Flickr30K test set \cite{young2014flickr30k}. Note the $x$-axis is in log-scale.}
    \label{fig:performance_comparison}
    \vspace{-10pt}
\end{figure}

In this work, our key insight is that off-the-shelf unimodal encoders that have been pre-trained on large amounts of unimodal data already encode rich semantics that should provide an effective bootstrap for multimodal fusion. We introduce FuseMix, a simple and easy-to-implement data augmentation scheme for multimodal fusion inspired by mixup \cite{zhang2018mixup}, where we share the mixing coefficient across modalities. We show that by aligning the latent spaces of existing pre-trained unimodal encoders using FuseMix, we obtain highly competitive fused multimodal models, which in certain cases even outperform state-of-the-art methods in both image-text and audio-text retrieval tasks, all while using orders of magnitude less compute and data. For example, we use $\sim \! 600\times$ less compute  ($\sim \! 5$\footnote{To pre-compute 5M latent encodings for the pre-trained image and text encoders in our experiments, we require up to $\sim \! 4$ days, noting that this is a one-time procedure whose cost can be amortized. Then we need $\sim \! 1$ day to perform FuseMix fusion on the resulting latents, all using 1 V100 GPU, for a total of $\approx \! 5$ GPU days.  See Sec. \ref{sec:experiments]} for details.} vs. $\sim \! 3000$\footnote{CLIP trained for $\sim \! 12$ days on $256$ V100 GPUs $\approx \! 3072$ GPU days.}  GPU days) and $\sim \! 80\times$ less image-text pairs ($\sim \! 5$M vs. $\sim \! 400$M) than CLIP \cite{radford2021clip} to perform multimodal fusion, yet are still able to outperform it in recall for the text-to-image retrieval task on the Flickr30K test set \cite{young2014flickr30k}, see \autoref{fig:performance_comparison}. Moreover, in settings with access to limited multimodal pairs, we show that dataset quality and diversity are important properties to increase downstream performance. Finally, we further demonstrate the applicability of our FuseMix fusion framework for audio-to-image generation \cite{girdhar2023imagebind}.

%% file: sections/2_related.tex
\section{Related Work}
\label{sec:related}

\textbf{Multimodal Learning.} The overarching objective of multimodal learning is to build universal models that can jointly perceive data of various modalities \cite{tan2019lxmert, li2020oscar, yuan2021florence, wang2022internvideo, bao2022vlmo, wang2022simvlm, girdhar2022omnivore, likhosherstov2023polyvit, zhang2023metatransformer, wu2023next}. Said modalities can range from data streams including but not limited to image, text, audio, and video. A standard approach to building multimodal models is to train them end-to-end on data paired across all modalities of interest \cite{arandjelovic2017looklistenlearn, lu2019vilbert, sun2019videobert, su2020vlbert, chen2020uniter, li2021albef, li2022blip}. However, this approach generally does not scale well since training large-scale multimodal models from scratch can quickly become very compute and data intensive.
A more practical approach is to instead bootstrap from pre-trained unimodal networks. Yet, several works in this vein still perform backpropagation through the pre-trained networks \cite{tsimpoukelli2021frozen, alayrac2022flamingo, li2023blip2, chen2023pali, liu2023llava,  driess2023palme, moon2023anymal}, which incurs significant overhead due to the large size of the underlying unimodal networks; this problem is bound to be exacerbated as the size of networks increases.

More related to our setting are multimodal models that focus on learning a single shared latent space wherein multiple modalities can be jointly encoded (i.e.\ multimodal alignment). This line of work was pioneered by CLIP \cite{radford2021clip} and ALIGN \cite{jia2021align}, which use a dual-encoder architecture trained with a contrastive objective to jointly embed texts and images. CoCa \cite{yu2022coca} adds an autoregressive image captioning term to the contrastive objective, which they find improves performance. 3T \cite{kossen2023three} instead aligns the text and image encoders with the latent space of a pre-trained classifier. LiT \cite{zhai2022lit} uses a frozen pre-trained image classifier as the image encoder, and aligns a text encoder with it. Despite their successes, all of these works train one or both encoders from scratch, requiring expensive gradient computations spanning many GPUs. They also use internet-scale datasets consisting of image-text pairs ranging in quantity from 400M to 5B pairs, and these datasets are often not made publicly available. Moreover, several works have extended CLIP to include other modalities such as video \cite{gorti2022xpool, luo2022clip4clip, fang2021clip2video} and audio \cite{guzhov2022audioclip}, but they require fine-tuning CLIP to achieve good performance. Similarly, other audio-text fusion methods \cite{deshmukh2022audio, laionclap2023, mei2022} require fine-tuning of the underlying encoders and additional training data. Finally, ImageBind \cite{girdhar2023imagebind} learns a shared latent space across six modalities using a contrastive objective with images as an anchor modality, which they achieve by jointly training several modality encoders from scratch. In contrast to all these works, we prioritize computational and data efficiency by using frozen pre-trained unimodal encoders, by leveraging minimal multimodal paired data, and by ensuring all our experiments require no more than a single GPU of compute.   

\textbf{Data Augmentation.}
Historically, data augmentations were introduced in an effort to synthetically increase dataset size and diversity \cite{lecun1998, simonyan2015}: this is exactly our goal, as we operate in a setting with relatively scarce paired multimodal data. In the natural image domain, common augmentations include horizontal flips, random crops, and color jitter \cite{chen2020simclr, balestriero2023cookbook}, which were designed to leave semantic information unchanged. However, designing such augmentations in any given domain requires expert knowledge of which transformations preserve semantic information. For example, na\"ively applying color jitter on the medical image domain can destroy the most relevant information for tasks like cancer classification \cite{sui2023lfr, shen2022randstainna}. Furthermore, handcrafted augmentation schemes typically do not readily transfer to other modalities. This effect is evidenced by the scarcity of modality-agnostic augmentation schemes, despite recent efforts therein such as random projections \cite{sui2023lfr} and randomized quantization \cite{wu2023randomizedquantization}. We note that while input masking has been successfully applied in various modalities, expert knowledge is still required to determine an appropriate masking strategy for each modality individually \cite{devlin2018bert, he2022masked, tong2022videomae, huang2022masked}. Given these challenges, it is unsurprising that data augmentations are not as well studied for multimodal learning \cite{hao2023mixgen}. In our work, we propose a multimodal augmentation scheme that operates on latent space and is inspired by mixup \cite{zhang2018mixup}.

%% file: sections/3_background.tex
\section{Problem Setting and Background}
\label{sec:background}

 \subsection{Multimodal Fusion as Alignment}
In this work, we define multimodal fusion from the perspective of alignment. Alignment is the task of learning a single latent space that is shared between multimodal inputs. Formally, given any two data modalities $\mathcal{X}$ and $\mathcal{Y}$ (e.g. images and texts), we aim to learn two networks, $f_X \colon \mathcal{X} \to \mathcal{S}$ and $f_Y \colon \mathcal{Y} \to \mathcal{S}$, that embed each respective modality into a shared latent space $\mathcal{S}$.

Recently, contrastive learning has emerged as a prevalent objective for multimodal alignment \cite{radford2021clip, jia2021align, li2021albef, zhai2022lit}. It aims to learn a joint latent space wherein semantically similar multimodal inputs in ambient space are encoded to nearby points, while semantically dissimilar inputs are embedded further apart. To this end, contrastive learning requires access to semantically similar multimodal inputs in the form of positive pairs (e.g. images and their corresponding text captions), as well as access to semantically dissimilar negative pairs (e.g. unrelated images and texts). Therefore, we must assume there is a way to obtain samples of positive pairs from the joint distribution over modalities $\mathcal{X}$ and $\mathcal{Y}$ given by $p_{X,Y}$. Negative pairs are often obtained by sampling from the product of marginal distributions of each modality, $p_X$ and $p_Y$.\footnote{While sampling independently from the marginals could technically result in a semantically related (positive) pair, the probability of this happening in practice is extremely small (e.g.\ a random text describing an image will rarely properly describe a different random image), thus justifying this procedure to obtain negative pairs.} With access to a positive pair $(x, y) \sim p_{X, Y}$ and negative pairs $(x_i^-, y_i^-) \overset{\text{i.i.d.}}{\sim} p_X p_Y$ for $i=1,\dots, M$, contrastive learning  in the context of multimodal alignment leverages the InfoNCE loss \cite{vandenoord2018infonce}:
\begin{align}\label{eq:infonce}
     \mathcal{L} & \left(f_X, f_Y; x, y, \{y_i^-\}_{i=1}^M\right) \triangleq \\
    & \hspace{5pt} -\log \dfrac{e^{f_X(x) \cdot f_Y(y)/\tau}}{e^{f_X(x) \cdot f_{Y}(y)/\tau}{+} \sum_{i=1}^M e^{f_X(x) \cdot f_Y(y_i^-)/\tau}}, \nonumber
\end{align}
where $a \cdot b \triangleq \frac{a^\top b}{\Vert a\Vert_2 \Vert b\Vert_2}$ denotes cosine similarity\footnote{We slightly abuse notation here and denote cosine similarity using the commonly used dot product notation for conciseness.} and $\tau>0$ is either a fixed or learnable scalar temperature parameter. The final objective is then given by a symmetric version \cite{radford2021clip, jia2021align} of the InfoNCE objective:
\begin{align}\label{eq:sym}
    \mathcal{L}_{\text{sym}}\left(f_X, f_Y\right) \triangleq \mathbb{E}\Big[&\tfrac{1}{2}\mathcal{L} \left(f_X, f_Y; x, y, \{y_i^-\}_{i=1}^M \right) \\
    & + \tfrac{1}{2}\mathcal{L} \left(f_Y, f_X;y, x, \{x_i^-\}_{i=1}^M \right) \Big], \nonumber
\end{align}
where the expectation is taken with respect to the positive pair $(x, y) \sim p_{X, Y}$ and the $M$ negative pairs $(x_i^-, y_i^-) \overset{\text{i.i.d.}}{\sim} p_X p_Y$. 

We note that formulating alignment through contrastive learning has been shown to enable zero-shot transfer to various multimodal downstream tasks \cite{radford2021clip, girdhar2023imagebind, guzhov2022audioclip, liu2023tr0n}, and has also been shown to improve performance in general multimodal settings, including understanding-based \cite{li2021albef, li2022blip} and generation-based \cite{li2023blip2, dai2023instructblip, li2023blipdiffusion} tasks. This formulation also admits theoretical motivations from the perspective of mutual information maximization \cite{vandenoord2018infonce, bachman2019learning, tian2020contrastive, li2021albef}.

\subsection{Mixup}
Mixup \cite{zhang2018mixup} is a general-purpose data augmentation routine for supervised learning. Its premise is simple: given pairs $(x, l)$ and $(\hat{x}, \hat{l})$ of data (i.e.\ $x$ and $\hat{x}$) and their corresponding labels (i.e.\ $l$ and $\hat{l}$), it constructs augmented samples by taking the convex combinations $\tilde{x} \triangleq \lambda x + (1-\lambda) \hat{x}$ and $\tilde{l} \triangleq \lambda l + (1-\lambda) \hat{l}$, where $\lambda \in (0,1)$ is an interpolation coefficient most commonly sampled from a Beta distribution $\mathcal{B}(\alpha, \beta)$ with hyperparameters $\alpha, \beta > 0$. The loss used to train the model is then optimized on the augmented data/label pairs rather than the original ones. Subsequent works have motivated mixup from the perspective of robustness and generalization \cite{zhang2020does}, as well as calibration \cite{thulasidasan2019mixup}. Variations on mixup have extended the method to contrastive learning where labels are unavailable \cite{verma2021dacl}, but can be created by proxy \cite{lee2021imix}. Recently, in the context of multimodal learning, \citet{so2022geodesic} proposed a mixup strategy using spherical interpolations to fine-tune CLIP, but this method requires a shared latent space that is already aligned and is not readily applicable in our setting.

%% file: sections/4_motivation.tex
\section{Motivation}
\label{sec:motivation}
\begin{figure*}
    \centering
     \includegraphics[scale=0.171, trim={2 43 10 90}, clip]{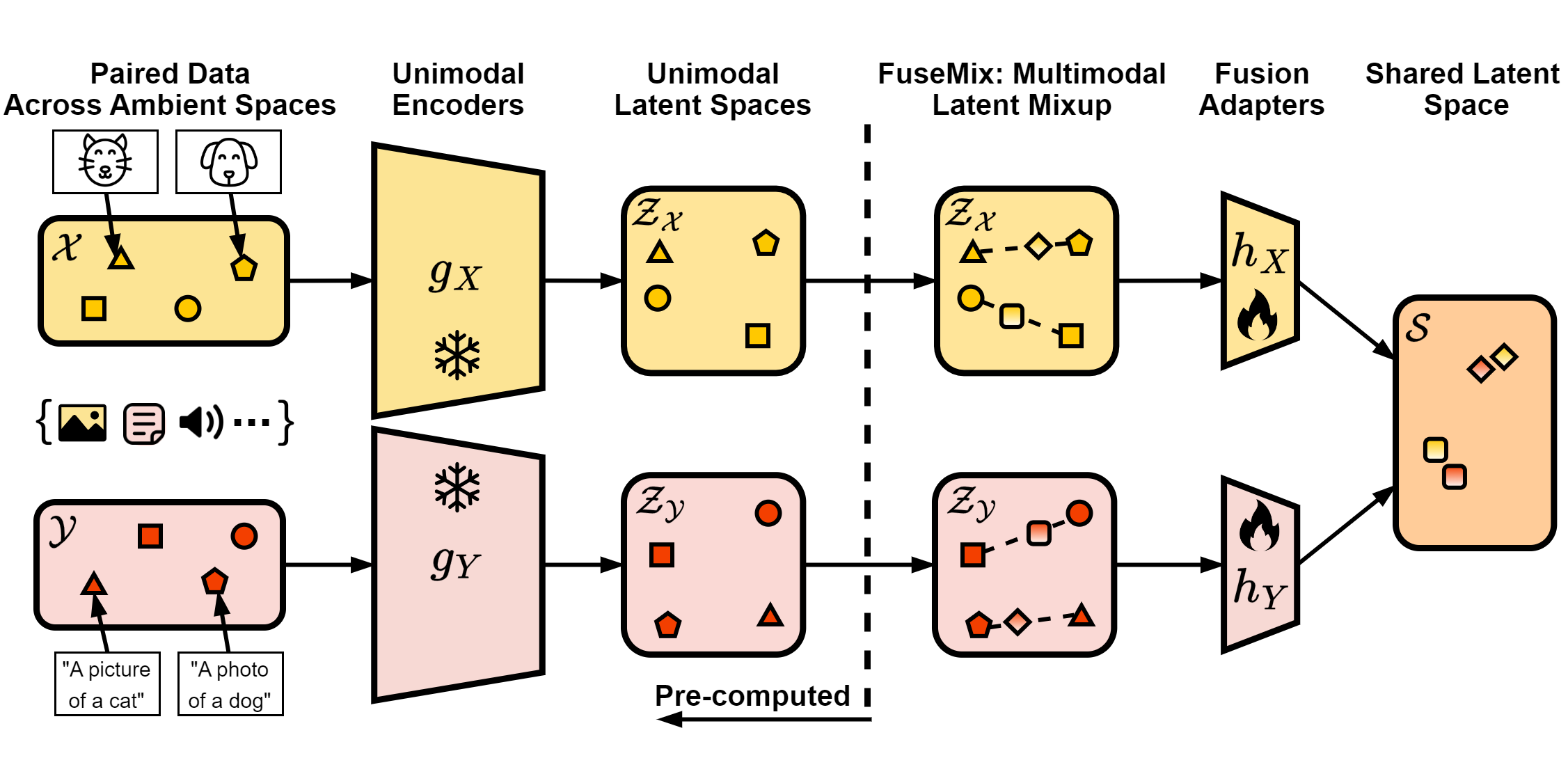}
    \caption{A schematic of our proposed fusion framework to align the latent spaces of pre-trained unimodal encoders using a minimal set of paired data. The unimodal encoders are kept frozen, and their latent encodings are pre-computed only once. FuseMix applies mixup on each latent space, importantly sharing the mixing coefficient across modalities, and is used as a modality-agnostic data augmentation. Then, the lightweight fusion adapters are trained to align the resulting augmented latents into a shared latent space.}
    \label{fig:illustration}
    \vspace{-8pt}
\end{figure*}

Despite recent successes, the prototypical paradigm for multimodal fusion exhibits critical bottlenecks rooted in large computational and data overhead, as well as a lack of modularity. In this section, we discuss these bottlenecks:

\textbf{Computational Burden.} Recent advances in deep learning have shown that model scale is a key driver of performance and downstream capabilities \cite{kaplan2020scaling, brown2020gpt3, zhai2022scalingvit, liu2022convnext, dehghani2023vit22b}. 
Although increasing model scale can greatly benefit performance, the required computational cost to train such models also increases commensurately, and is unattainable for many machine learning practitioners and researchers. In the context of multimodal models, these effects are more prominent as computational requirements are generally compounded.  For example, in our setting of multimodal fusion, it is common to jointly train both $f_X$ and $f_Y$ \cite{radford2021clip, jia2021align, yu2022coca, girdhar2023imagebind, kossen2023three}. This means that backpropagation is now required through two networks that must both be held in memory. Moreover, as we increase the scale of each network, the number of parameters requiring expensive gradient computations quickly accumulates. We therefore aim to prioritize computational considerations to design an efficient framework for multimodal fusion.

\textbf{Scarcity of High-Quality Paired Data.} Sourcing multimodal paired data is a necessary step in most multimodal applications. This step amounts to obtaining paired samples from the joint distribution over modalities as  $(x, y) \sim p_{X,Y}$. However, in practice, high-quality paired data is often scarce and expensive to obtain. Typically, this is either due to a lack of readily available paired data across all modalities of interest \cite{girdhar2023imagebind}, or due to noisy samples stemming from large amounts of uninformative and weakly-labeled data pairs \cite{li2022blip, wang2023too}. On the other hand, high-quality samples of unimodal data from the corresponding marginal distributions of each modality, $x \sim p_X$ and $y \sim p_Y$, are relatively cheap and easy to amass in large quantities. This is because unimodal data can be collected without any label pairings while still providing informative intrinsic supervisory signals, as evidenced by successes in self-supervised learning \cite{devlin2018bert, he2020momentum, chen2020simclr, he2022masked, balestriero2023cookbook}. As such, we aim to defray the cost of sourcing multimodal paired data by leveraging more readily available unimodal signals.

\textbf{Tight Coupling From End-to-End Fusion.}
While jointly training $f_X$ and $f_Y$ from scratch for multimodal fusion may produce a semantically meaningful shared latent space, the resulting networks are tightly coupled. This means that modifying any aspect of either network typically requires completely re-training both networks end-to-end. This presents a challenging bottleneck in practice, as research advancements in each underlying modality cannot be incorporated by end-to-end multimodal fusion without re-training $f_X$ and $f_Y$, incurring significant computational, data, and environmental costs \cite{strubell2019energy}. Our goal is therefore to design a plug-and-play framework for multimodal fusion such that individual components can be easily replaced with minimal overhead, allowing multimodal models to keep pace with unimodal improvements. 

%% file: sections/5_method.tex
\section{Method}
\label{sec:method}
In this section we present our framework for multimodal fusion, which aims to address the key considerations of computational and data efficiency, as well as modularity (Sec. \ref{sec:method-eff}). We also introduce a multimodal augmentation scheme on latent space called FuseMix to facilitate multimodal fusion (Sec. \ref{sec:method-multimix}). Our entire pipeline is depicted in \autoref{fig:illustration}.

\subsection{Towards Efficient Multimodal Fusion}
\label{sec:method-eff}
 As a first step, we take our two encoders as $f_X=h_X \circ g_X$ and $f_Y=h_Y \circ g_Y$. That is, we define $g_X \colon \mathcal{X} \to \mathcal{Z_X}$ and $g_Y \colon \mathcal{Y} \to \mathcal{Z_Y}$, where $\mathcal{Z_X}$ and $\mathcal{Z_Y}$ are intermediate latent spaces. We then have $h_X \colon \mathcal{Z_X} \to \mathcal{S}$ and $h_Y \colon \mathcal{Z_Y} \to \mathcal{S}$, which we hereafter refer to as fusion adapters. Our key insight here is to take both $g_X$ and $g_Y$ as pre-trained unimodal encoders which we keep frozen throughout, and treat our fusion adapters $h_X$ and $h_Y$ as learnable heads for multimodal fusion. This design offers several advantages:

\textbf{Computational Improvements.}  We can now equivalently rewrite the alignment loss from \autoref{eq:infonce} as $\mathcal{L}(h_X, h_Y; g_X(x), g_Y(y), \{g_Y(y_i^-)\}_{i=1}^M)$.
This allows us to express the contrastive objective in \autoref{eq:sym} as an expectation with respect to: positive pairs of encodings $(g_X(x), g_Y(y))$, whose distribution is induced by pushing positive pairs $(x, y) \sim p_{X, Y}$ through the encoders $g_X$ and $g_Y$; and negative pairs of encodings $(g_X(x_i^-), g_Y(y_i^-))$, whose distribution is analogously obtained from negative pairs on the ambient spaces. 
Importantly, since this expectation is taken with respect to a distribution which depends only on the frozen $g_X$ and $g_Y$, but not on the trainable $h_X$ and $h_Y$, the unimodal encoders $g_X$ and $g_Y$ are not used in any gradient computations. In other words, since the unimodal encoders are only needed to provide samples on latent space, not for backpropagation, we can simply pre-compute these samples and then discard the unimodal encoders while training. 
This step ensures that we do not need to store large encoders in memory during multimodal fusion, which significantly reduces computational requirements. The only parameters stored in memory during fusion are those of the learnable fusion adapters which are 
extremely lightweight compared to the unimodal encoders. In fact, in all of our experiments, we only require a single GPU at every step.

\textbf{Paired Data Efficiency.} By setting $\mathcal{Z_X}$ and $\mathcal{Z_Y}$ as the latent spaces of pre-trained unimodal encoders, we can directly benefit from the rich modality-specific semantics that they already encode. Learning this information from scratch might be redundant for multimodal fusion, so leveraging pre-trained unimodal encoders can be an effective bootstrap to reduce the need for large-scale multimodal paired data. We can interpret this effect as a form of distillation from a unimodal latent space into a joint space for which contrastive objectives have been shown to be effective \cite{tian2019contrastive, kossen2023three}. In other words, leveraging pre-trained unimodal encoders for multimodal fusion should require less paired data than training end-to-end from scratch.

\textbf{Plug-and-Play Framework.} We highlight that our modular approach to multimodal fusion is agnostic to both the choice of unimodal encoders $g_X$ and $g_Y$ and to the underlying modalities $\mathcal{X}$ and $\mathcal{Y}$. Importantly, by combining arbitrary pre-trained unimodal encoders, we can decouple unimodal learning from multimodal fusion. Therefore, as the development of unimodal encoders continues to advance, we can easily and efficiently leverage new unimodal encoders for multimodal fusion in a plug-and-play manner.

\subsection{FuseMix: Multimodal Latent Mixup}
\label{sec:method-multimix}
Given our aim of performing multimodal fusion with minimal samples of paired data, it would seem intuitive to also leverage data augmentations to generate synthetic  multimodal pairs $(\tilde{x}, \tilde{y}) \in \mathcal{X} \times \mathcal{Y}$. 
However, constructing semantically meaningful data augmentations directly on the ambient spaces $\mathcal{X}$ and $\mathcal{Y}$ is generally challenging due to the heterogeneity of multimodal data \cite{hao2023mixgen}. On the other hand, we note that $\mathcal{Z_X}$ and $\mathcal{Z_Y}$ provide a more homogeneous alternative since they are both intermediate latent spaces of pre-trained unimodal encoders.  Additionally, they already encode semantic information that can be beneficial for creating meaningful data augmentations. 

As such, we introduce a simple yet effective multimodal augmentation scheme on latent space that is agnostic to both the involved modalities and the choice of unimodal encoders. Our approach, which we call FuseMix, is inspired by mixup \cite{zhang2018mixup}, in that augmented samples are generated from random convex combinations. In particular, we take linear interpolations between samples in both $\mathcal{Z_X}$ and $\mathcal{Z_Y}$. Importantly, since both latent spaces are taken from pre-trained unimodal encoders, we should expect linear interpolations to be more semantically meaningful than when carried out on ambient space, as is typically done in mixup \cite{zhang2018mixup, verma2021dacl, lee2021imix}. We note that this idea of semantic interpolations in latent space is reminiscent of latent space arithmetic that has a well-established history \cite{mikolov2013efficient, pennington2014glove, ethayarajh2018towards, girdhar2023imagebind}. 

However, na\"ively mixing random samples in each latent space would only produce augmented pairs of latents $(\tilde{z}_x, \tilde{z}_y) \in \mathcal{Z_X} \times \mathcal{Z_Y}$ where $\tilde{z}_x$ and $\tilde{z}_y$ are unrelated to one another. Therefore, we want to impose some structure on how interpolations are performed across modalities to ensure that we can construct semantically meaningful augmented pairs. To achieve this we take any two existing positive multimodal pairs $(z_x, z_y) \triangleq (g_X(x), g_Y(y))$ and $(\hat{z}_x, \hat{z}_y) \triangleq (g_X(\hat{x}), g_Y(\hat{y}))$, where $(x, y), (\hat{x}, \hat{y}) \overset{\text{i.i.d.}}{\sim} p_{X,Y}$, and construct a corresponding augmentation $(\tilde{z}_x, \tilde{z}_y)$ as
\begin{equation}
\left(\tilde{z}_x, \tilde{z}_y \right) \triangleq \lambda \left( z_x, z_y \right) + (1 - \lambda) \left( \hat{z}_x, \hat{z}_y \right),
\end{equation}
where $\lambda \in (0,1)$ is the shared interpolation coefficient. Sharing $\lambda$ across modalities ensures that the resulting augmentation is semantically consistent, meaning $\tilde{z}_x$ and $\tilde{z}_y$ still form a valid positive pair. 
In practice, we can of course similarly apply FuseMix to obtain interpolations of negative pairs in such a way that the result remains a negative pair.
Finally, our version of \autoref{eq:sym} on the intermediate latent spaces with FuseMix is given by
\begin{align}
\text{ \small
    $\mathcal{L}_{\text{sym}}^{\text{FuseMix}}\left(h_X, h_Y\right) \triangleq \mathbb{E}\Big[$} & \text{ \small $\tfrac{1}{2} \mathcal{L}\left(h_X, h_Y; \tilde{z}_x, \tilde{z}_y, \{\tilde{z}_{y_i}^-\}_{i=1}^M\right)$}\\
    & \text{ \small $+ \tfrac{1}{2} \mathcal{L}\left(h_Y, h_X; \tilde{z}_y, \tilde{z}_x, \{\tilde{z}_{x_i}^-\}_{i=1}^M\right)\Big]$},\nonumber
\end{align}
where the expectation is taken with respect to: the positive pairs $(z_x, z_y)$ and $ (\hat{z}_x, \hat{z}_y)$ used to obtain the augmented positive pair $(\tilde{z}_x, \tilde{z}_y)$; the negative pairs $\{(z_{x_i}^-, z_{y_i}^-)\}_{i=1}^M$ and $\{(\hat{z}_{x_i}^-, \hat{z}_{y_i}^-)\}_{i=1}^M$ used to obtain the augmented negative pairs $\{(\tilde{z}_{x_i}^-, \tilde{z}_{y_i}^-)\}_{i=1}^M$; and $\lambda \sim \mathcal{B}(\alpha, \beta)$. We note that our FuseMix fusion algorithm can be implemented very easily. Given pre-computed samples of multimodal latent pairs (see Sec. \ref{sec:method-eff}), setting the batch size $B \triangleq M + 1$, and taking $\alpha=\beta$, the simplicity of our method is illustrated in Algorithm \ref{algo:mixup}, requiring only a few lines of code.

\makeatletter
\patchcmd{\@algocf@start}
  {-1.5em}
  {2pt}
  {}{}
\makeatother

\DecMargin{1.1em}
\begin{algorithm}[t]
\scriptsize
\SetAlCapNameFnt{\footnotesize}
\SetAlCapFnt{\footnotesize}
\caption{PyTorch-style pseudocode for FuseMix fusion.}
\label{algo:mixup}
\SetAlgoLined
    \PyComment{h\_X, h\_Y: learnable fusion adapters} \\
    \PyComment{B: batch size} \\
    \PyComment{D\_x, D\_y: latent dimension of unimodal encoders} \\
    \PyComment{D\_s: latent dimension of shared space} \\
    \PyComment{alpha: mixup Beta distribution hyperparameter} \\
    \PyComment{t: learnable temperature parameter} \\
    \BlankLine
    \PyComment{load latent pairs of batch size 2B} \\
    \PyCode{for z\_x,z\_y in loader:} \PyComment{(2B x D\_x, 2B x D\_y)}\\
    \Indp   
         \PyComment{FuseMix} \\
         \PyCode{z\_x1, z\_x2 = torch.chunk(z\_x, 2)} \PyComment{B x D\_x}\\ 
         \PyCode{z\_y1, z\_y2 = torch.chunk(z\_y, 2)} \PyComment{B x D\_y}\\ 
         \PyCode{lam = random.beta(alpha, alpha)} \\ 
         \PyCode{z\_x = lam * z\_x1 + (1 - lam) * z\_x2} \\ 
         \PyCode{z\_y = lam * z\_y1 + (1 - lam) * z\_y2} \\
         \BlankLine
         \PyComment{joint space and normalize} \\
         \PyCode{s\_x = l2\_normalize(h\_X(z\_x), dim=1)} \PyComment{B x D\_s} \\ 
         \PyCode{s\_y = l2\_normalize(h\_Y(z\_y), dim=1)} \PyComment{B x D\_s} \\ 
         \BlankLine
         \PyComment{pairwise cosine similarity w/ temperature} \\
         \PyCode{logits\_xy = (s\_x @ s\_y.T) * t.exp()} \PyComment{B x B} \\
         \PyCode{logits\_yx = (s\_y @ s\_x.T) * t.exp()} \PyComment{B x B} \\
         \BlankLine
         \PyComment{symmetric alignment loss} \\
         \PyCode{labels = torch.arange(B)} \\ 
         \PyCode{loss\_xy = cross\_entropy\_loss(logits\_xy, labels)} \\ 
         \PyCode{loss\_yx = cross\_entropy\_loss(logits\_yx, labels)} \\ 
         \PyCode{loss = (loss\_xy + loss\_yx) / 2} \\ 
         \BlankLine
         \PyComment{optimize} \\
         \PyCode{optimizer.zero\_grad()} \\ 
         \PyCode{loss.backward()} \\ 
         \PyCode{optimizer.step()} \\ 
    \Indm 
\end{algorithm}
\setlength{\textfloatsep}{3pt}

%% file: sections/6_experiments.tex
\section{Experiments}
\label{sec:experiments]}
In our experiments, we consider the image-text and audio-text modality pairings. We start by describing details of our implementation and then we perform experimental analysis to evaluate our framework and provide insights on key components of multimodal fusion.    

\subsection{Implementation Details}
\label{sec:impl}
\textbf{Unimodal Latent Extraction.} Since an important consideration of our method is to minimize computational requirements, we only use a single 32GB NVIDIA V100 GPU for all of our experiments. This is possible for us because, as mentioned in Sec. \ref{sec:method-eff}, we can pre-compute the latents from pre-trained unimodal encoders so that the underlying encoders can be discarded thereafter. Additionally, we can extract the latents for each modality one at a time to ensure that no more than one encoder must be loaded at once. Importantly, these steps allow us to consider large-scale encoders on the order of billions of parameters which would generally not be feasible for end-to-end fusion on a single GPU. We mainly consider Transformer-based \cite{vaswani2017attention} unimodal encoders, and extract low-dimensional latents from the penultimate layer of either the \texttt{[CLS]} token if it exists, or the mean-pooled token otherwise.  

\textbf{Multimodal Latent Fusion.}
We parameterize our fusion adapters as lightweight MLPs using an inverted bottleneck architecture following previous work \cite{lin2015far, tolstikhin2021mlp, bachmann2023scaling}. Each MLP consists of residual blocks followed by a final projection layer of dimension 512 by default to embed each modality into a shared space. We highlight that since our fusion adapters are operating on low-dimensional latents, the computational cost to train them is minimal, and despite training on a single GPU, we can use large batch sizes (up to $B=20$K on our V100 GPU), which has been shown to benefit contrastive learning \cite{wu2018unsupervised, tian2020contrastive, he2020momentum, chen2020simclr, wang2020understanding}. Finally, we note that in all of our experiments, unless otherwise stated, we use $\mathcal{L}_\text{sym}^\text{FuseMix}$ as our sole objective for multimodal fusion. More details on the MLP architecture and hyperparameters can be found in Appendix \ref{sec:appendix-arch} and \ref{sec:appendix-impl}.

\textbf{Training Datasets.} We rely on common multimodal datasets for training. Specifically, following previous works \cite{chen2020uniter, li2021albef, li2022blip, li2023blip2}, we leverage the image-text pairs from human-annotated datasets (COCO \cite{lin2014mscoco} and Visual Genome \cite{krishna2017visualgenome}), and web datasets (SBU Captions \cite{ordonez2011sbucaptions} and Conceptual Captions 3M \cite{sharma2018conceptualcaptions}), amounting to 5M total pairs. In order to remain data-efficient, we note that we intentionally avoid internet-scale datasets like the ones used in several recent works \cite{radford2021clip, jia2021align, yu2022coca, kossen2023three}, as these are orders of magnitude larger than our collated dataset. Similarly, to remain data-efficient for the audio-text regime, we only leverage the AudioCaps \cite{audiocaps} and Clotho \cite{clotho} train sets which provide 50K and 15K human-annotated audio-text pairs, respectively.

\subsection{Cross-Modal Retrieval Performance}
\label{sec:retrieval}

To assess the quality of multimodal alignment learned from FuseMix fusion, we follow previous works \cite{radford2021clip, jia2021align, kossen2023three, koepke22audio, deshmukh2022audio, laionclap2023} and evaluate our method using the downstream task of cross-modal retrieval. In particular, for the image-text pairing, we evaluate downstream performance on the Flickr30K \cite{young2014flickr30k} and COCO \cite{lin2014mscoco} test sets, and for the audio-text pairing, we evaluate our method on the AudioCaps \cite{audiocaps} and Clotho \cite{clotho} test sets. In our experiments, we use subscripts to specify which pre-trained unimodal encoders were used for bootstrapping. In terms of image encoders, we consider both DINOv2 \cite{oquab2023dinov2} (D) and UNICOM \cite{an2023unicom} (U) since, as of the time of writing, they are two of the top-ranked visual recognition models as measured by the ImageNet \cite{deng2009imagenet} linear probing benchmark. On the text side, we use the MTEB \cite{muennighoff2023mteb} text embedding benchmark to select two encoders with demonstrably semantic latent spaces, namely BGE \cite{xiao2023c} (B) and E5 \cite{wang2022e5} (E). Finally, on the audio side we utilize the commonly used HTS-AT \cite{htsatke2022} (H) and the recent Whisper \cite{radford2023whisper} (W) encoders. In practice, we actually use the concatenation of the latents from these two encoders (W\&H), similar to \cite{deshmukh2022audio}. We emphasize that given the plug-and-play nature of our method, as better unimodal encoders become available, we can quickly and cheaply incorporate them into our framework. We report results across all combinations of these encoders in \autoref{tab:text-image-ret} and \autoref{tab:text-audio-ret}.

For image-text retrieval, we highlight that our method is highly competitive and sometimes able to outperform various state-of-the-art methods which are trained on orders of magnitude more paired data and that require substantially more than a single GPU of compute for fusion. Moreover, we find that the combination of two of the most recent models, DINOv2+BGE, achieves the highest performance, highlighting the benefits of a plug-and-play approach that can leverage the most recent advancements. We also note that when our method and CLIP \cite{wang2023too} are both only trained on pairs from Conceptual Captions 3M, we outperform CLIP by a notable margin, demonstrating that FuseMix is an effective strategy for fusion on lower data regimes. Similarly, for audio-text retrieval we outperform all other methods trained on similar data, and can compete with methods that use orders of magnitude more paired data.

\begin{table*}[t]
    \centering
    \footnotesize
    \begin{tabular}{cccccccccccccc}
        \hline
        &  & \multicolumn{6}{c}{Flickr30K (1K test set)} & \multicolumn{6}{c}{COCO (5K test set)} \\
        \cline{3-14}
        Method & \# (image, text) & \multicolumn{3}{c}{text $\rightarrow$ image} & \multicolumn{3}{c}{image $\rightarrow$ text} & \multicolumn{3}{c}{text $\rightarrow$ image} & \multicolumn{3}{c}{image $\rightarrow$ text} \\
        & & R@1 & R@5 & R@10 & R@1 & R@5 & R@10 & R@1 & R@5 & R@10 & R@1 & R@5 & R@10 \\
        \hline
        \textit{internet-scale}\\
        FILIP \cite{yao2021filip} & 300M & 75.0 & 93.4&  96.3 & \textbf{89.8} & \textbf{99.2} & \textbf{99.8} & 45.9&  \textbf{70.6} & \textbf{79.3} & 61.3 & \textbf{84.3} & \textbf{90.4}\\
        CLIP \cite{radford2021clip} & 400M & 68.7 & 90.6 & 95.2 &88.0 & 98.7 & 99.4 & 37.8 & 62.4 & 72.2 & 58.4 & 81.5 & 88.1 \\
        ALIGN \cite{jia2021align} & 1B & \textbf{75.7} & \textbf{93.8} & \textbf{96.8} &  88.6 & 98.7 & 99.7 &  45.6 & 69.8 & 78.6 & 58.6 & 83.0 & 89.7 \\
        LiT \cite{zhai2022lit} & 4B &  66.5& -&- &83.9  & -&- & 43.6 &- &- & 59.5 & -& - \\
        3T \cite{kossen2023three} & 5B & 72.1 & -& -& 87.3 & -& - &  \textbf{48.5} & - & - & \textbf{64.1} & -& - \\
        \hline
        \textit{low-data regime}\\
        CLIP \cite{wang2023too} & 3M &  54.3 & 84.1 & 90.8 & 67.4 & 83.2 & 92.4 &  29.9 & 57.9 & 66.9  & 36.2 & 64.3 & 80.1 \\
        $\text{FuseMix}_{\text{(D,B)}}$ & 3M &  59.9 & 86.4 & 91.6 & 74.4 & 94.0 & 97.4 &  32.2 & 58.2 & 69.4 & 42.3 & 68.4 & 78.9 \\
        $\text{FuseMix}_{\text{(U,B)}}$ & 5M & 66.3 & 88.9 & 93.3 & 81.2 & 95.9 & 97.7 & 42.5 & 70.2 & 80.0   & 59.1 & 83.4 & 90.3 \\
        $\text{FuseMix}_{\text{(U,E)}}$  & 5M & 64.3 &  87.7 & 93.0 &80.2 &95.6 & 98.1 & 42.9 & 70.0 & 80.1  & 59.1  & 83.9 & 91.0 \\
        $\text{FuseMix}_{\text{(D,E)}}$ & 5M & 68.8 & 90.9 & 94.6 & \textbf{85.2} & 96.9 & 98.4 & 46.1 & 74.3 &  \textbf{84.1} & \textbf{64.3} & 86.2 & 92.1 \\
        $\text{FuseMix}_{\text{(D,B)}}$ &  5M& \textbf{71.2} & \textbf{91.4} & \textbf{94.7}  & 84.8 & \textbf{97.2} & \textbf{99.1} & \textbf{46.3} & \textbf{74.6} &  83.4 & 62.7 & \textbf{86.4} & \textbf{92.7} \\
        \hline
    \end{tabular}
    \caption{Results of image-text retrieval on the Flickr30K 1K and COCO 5K test sets. The top section of the table contains fusion methods trained with internet-scale data, while the bottom section contains methods using much fewer image-text pairs. All our results use the largest available version of the underlying unimodal encoders. Refer to Sec. \ref{sec:retrieval} for the definition of the subscripts.}
    \label{tab:text-image-ret}
\end{table*}

\begin{table*}[t]
        \centering
        \footnotesize
	\begin{tabular}{cccccccccccccc}
		\hline
		   &  & \multicolumn{6}{c}{AudioCaps (1K test set)} & \multicolumn{6}{c}{Clotho (1K test set)} \\
		\cline{3-14}
            Method & \# (audio, text) & \multicolumn{3}{c}{text $\rightarrow$ audio} & \multicolumn{3}{c}{audio $\rightarrow$ text} & \multicolumn{3}{c}{text $\rightarrow$ audio} & \multicolumn{3}{c}{audio $\rightarrow$ text} \\
          & & R@1 & R@5 & R@10 & R@1 & R@5 & R@10 & R@1 & R@5 & R@10 & R@1 & R@5 & R@10 \\
          \hline
          \textit{internet-scale}\\
          LAION-CLAP \cite{laionclap2023} & 700K & 36.2 & 70.3 & 82.5 & 45.0 & 76.7 & 88.0 & 17.2 & 42.9 & 55.4 & 24.2 & 51.1 & \textbf{66.9} \\
          VALOR \cite{chen2023valor} & 6.5M &  40.1 & 73.9 & 83.1 & - & - & - & 17.5 & 42.7 & 55.3 & - & - & - \\
          ONE-PEACE \cite{wang2023one} & 2.4M & 42.5 & \textbf{77.5} & \textbf{88.4} & \textbf{51.0} & \textbf{81.9} & \textbf{92.0} & 22.4 & 49.0 & 62.7 & \textbf{27.1} & \textbf{52.3} & 65.4 \\
          VAST \cite{chen2023vast} & 27M & \textbf{52.0} & 76.8 & 82.9 & - & - & - & \textbf{26.9} & \textbf{53.2} & \textbf{66.1} & - & - & - \\
          \hline
          \textit{low-data regime}\\
          ML-ACT \cite{mei2022} & 50K/15K & 33.9 & 69.7 & 82.6 & 39.4 & 72.0 & 83.9 & 14.4 & 36.6 & 49.9 & 16.2 & 37.5 & 50.2 \\
          MMT \cite{koepke22audio} & 50K/15K & 36.1 & 72.0 & 84.5 & 39.6 & 76.8 & 86.7 & 6.5 & 21.6 & 32.8 & 6.3 & 22.8 & 33.3 \\
          CLAP-HTSAT \cite{deshmukh2022audio} & 70K & 34.7 & 70.2 & 82.0 & 41.9 & 73.2 & 84.6 & 16.8 & \textbf{41.1} & \textbf{54.1} & 20.0 & \textbf{44.9} & \textbf{58.7} \\
          LAION-CLAP \cite{laionclap2023} & 65K & 36.7 & 70.9 & 83.2 & 45.3 & 78.0 & 87.7 & 12.0 & 31.6 & 43.9 & 15.7 & 36.9 & 51.3 \\
          $\text{FuseMix}_{\text{(W\&H,B)}}$ & 50K/15K & 41.3 & 76.9 & \textbf{87.6} & 50.3 & 81.0 & 89.6 & 15.7 & 39.4 & 53.8 & 19.7 & 42.9 & 56.5 \\
          $\text{FuseMix}_{\text{(W\&H,B)}}$ & 65K & \textbf{43.1} & \textbf{77.4} & 87.4 & \textbf{52.4} & \textbf{83.4} & \textbf{92.4} & \textbf{17.6} & \textbf{41.1} & 53.5 & \textbf{21.5} & 44.7 & 57.4 \\
          \hline
	\end{tabular}
\caption{
Results of audio-text retrieval on the AudioCaps 1K and Clotho 1K test sets. The top section of the table contains fusion methods trained with internet-scale data, while the bottom section contains methods using much fewer audio-text pairs. `50K/15K` means that the model was trained only on the AudioCaps (50K) (resp. Clotho (15K)) training set when evaluating on AudioCaps (resp. Clotho). All our results use the largest available version of the underlying unimodal encoders. Refer to Sec. \ref{sec:retrieval} for the definition of the subscripts.}
\label{tab:text-audio-ret}
\vspace{-4pt}
\end{table*}

\subsection{Evaluating Dataset Efficiency}
\begin{figure}[htbp]
    \vspace{-8pt}
  \footnotesize
  \centering
  \begin{subfigure}[b]{3cm}
    \centering
    \hspace{-0.75cm}
    \begin{tikzpicture}
      \begin{axis}[
        width=3.5cm,
        height=3.5cm,
        xlabel={Subset Size (\%)},
        ylabel={R@1},
        xlabel style={yshift=0.2cm},
        ylabel style={yshift=-0.6cm},
        grid=major,
        ]
        \addplot[blue, mark=*, mark size=1] coordinates {(6.67,25.32) (13.33,31.88) (33.33,39.72) (67.67,44.1) (100,46.84)};
      \end{axis}
    \end{tikzpicture}
    \caption{Quantity}
    \label{fig:data-quantity}
  \end{subfigure}%
  \begin{subfigure}[b]{3cm}
    \centering
    \hspace{-1cm}
    \begin{tikzpicture}
      \begin{axis}[
        width=3.5cm,
        height=3.5cm,
        ybar,
        bar width=0.2cm,
        ylabel={R@1},
        nodes near coords={}, 
        nodes near coords align={vertical},
        xticklabels={
          {H \\ 500K},
          {W \\ 500K},
          {W \\ 3M}
        },
        xtick=data,
        xtick align=inside,
        x tick label style={
          align=center 
        },
        enlarge x limits=0.15,
        ylabel style={yshift=-0.65cm},
        ymin=40,
        grid=major,
        ]
        \addplot coordinates {(1,60) (2,51.14) (3, 59.9)};
      \end{axis}
    \end{tikzpicture}
    \caption{Quality}
    \label{fig:data-quality}
  \end{subfigure}%
  \begin{subfigure}[b]{3cm}
    \centering
    \hspace{-1cm}
    \begin{tikzpicture}
      \begin{axis}[
        width=3.5cm,
        height=3.5cm,
        xlabel={Subset Size (\%)},
        ylabel={$\Delta$ R@1 (\%)},
        xlabel style={yshift=0.2cm},
        ylabel style={yshift=-0.65cm},
        grid=major,
        ]
        \addplot[red, mark=*, mark size=1] coordinates {(6.67,38) (13.33,26.34) (33.33,12.48) (67.67,6.89) (100,0)};
      \end{axis}
      \vspace{-3pt}
    \end{tikzpicture}
    \caption{Diversity}
    \label{fig:data-diversity}
  \end{subfigure}%
  \vspace{-2pt}
  \caption{Measuring the effect of dataset quantity, quality, and diversity on downstream performance, evaluated using text-to-image retrieval on the Flickr30K test set. The $x$-axes indicate the relative/absolute number of image-text pairs, while H and W denote human and web-annotated, respectively. $\Delta$ R@1 (\%) denotes relative improvement in Recall@1 compared to uniform subsampling.}\label{fig:quantity-quality-diversity}
  \vspace{3pt}
\end{figure}
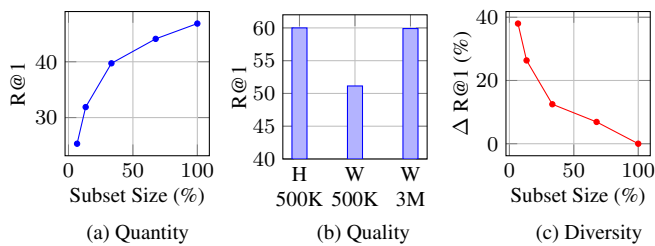

As mentioned in Sec. \ref{sec:motivation}, sourcing multimodal data pairs across all modalities of interest can be costly, especially in scarce data regimes. In practical settings, it is therefore natural to wonder how one should allocate efforts to construct a dataset for multimodal fusion that would maximize performance. We aim to answer this question by characterizing and quantifying three key properties of datasets, namely quantity, quality, and diversity. For dataset quantity, we take an existing dataset and uniformly subsample various numbers of pairs to measure the effect of quantity on downstream performance. For dataset quality, we consider human-annotated datasets to be of higher-quality, and web datasets to be of lower-quality. Finally, for dataset diversity, we can rely on determinantal point processes (DPPs) \cite{kulesza2011kdpp, kulesza2012determinantal, chen2018fast}. For a given dataset, DPPs return subsets of points of a pre-specified size that are maximally diverse (see Appendix \ref{sec:appendix-dpp} for details). Applied to our setting, we use DPPs on an existing dataset to obtain diverse subsets of various sizes and then compare the performance against uniformly sampled subsets of the corresponding sizes.


Our results are shown in \autoref{fig:quantity-quality-diversity}. We observe that increased quantity of data improves performance in lower data regimes as expected (\autoref{fig:data-quantity}). However, the quality of the underlying dataset also has a very strong effect, as has been similarly observed in other work \cite{li2022blip, wang2023too}. In fact, in \autoref{fig:data-quality}, we find that $6 \times$ the number of image-text pairs from the web are required to match the performance of using higher quality human-annotated pairs. Interestingly, in \autoref{fig:data-diversity} we find that with access to limited data, sourcing image-text pairs that are maximally diverse provides substantial improvements of up to nearly $40\%$ compared to selecting image-text pairs without consideration for diversity (i.e. uniform sampling). As such, when sourcing multimodal paired data in practice, it is important to consider not just quantity, but also quality and diversity, as these aspects can unlock notable improvements in scarce data regimes. 


\subsection{Audio-to-Image Generation}

\begin{figure} [t!]
\centering
\fontsize{6.8}{8}
\selectfont
\setlength{\tabcolsep}{3pt}
\begin{tabular}
{p{0.22\linewidth}p{0.22\linewidth}p{0.22\linewidth}p{0.22\linewidth}}

   \stackunder[5pt]{\includegraphics[scale=0.22]{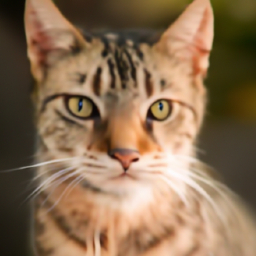}}
   {\href{https://github.com/layer6ai-labs/fusemix/blob/files/files/cat.wav}{\includegraphics[width=0.26cm, trim={0 12 0 12}, clip]{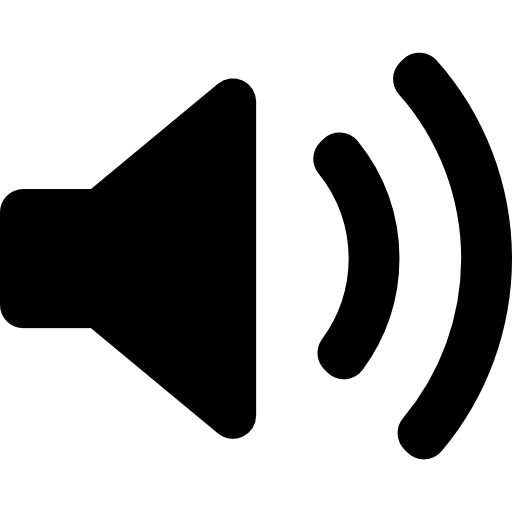}}}\vspace{5pt} & 
   \stackunder[5pt]{\includegraphics[scale=0.22]{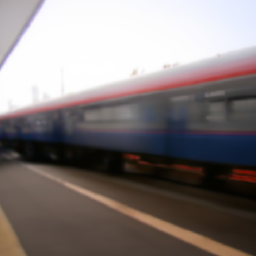}}
   {\href{https://github.com/layer6ai-labs/fusemix/blob/files/files/train.wav}{\includegraphics[width=0.26cm, trim={0 12 0 12}, clip]{images/audio-icon.png}}\vspace{5pt}}  & 
   \stackunder[5pt]{\includegraphics[scale=0.22]{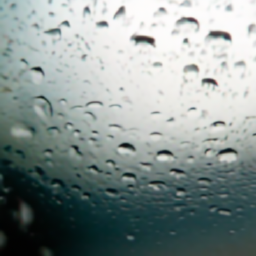}}
   {\href{https://github.com/layer6ai-labs/fusemix/blob/files/files/rain.wav}{\includegraphics[width=0.26cm, trim={0 12 0 12}, clip]{images/audio-icon.png}}\vspace{5pt}}  & 
   \stackunder[5pt]{\includegraphics[scale=0.22]{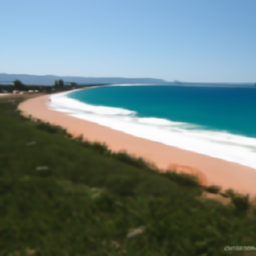}}
   {\href{https://github.com/layer6ai-labs/fusemix/blob/files/files/sea.wav}{\includegraphics[width=0.26cm, trim={0 12 0 12}, clip]{images/audio-icon.png}}\vspace{5pt}} \\
    \includegraphics[scale=0.22]{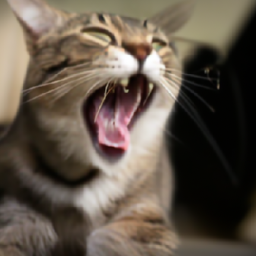} & \includegraphics[scale=0.22]{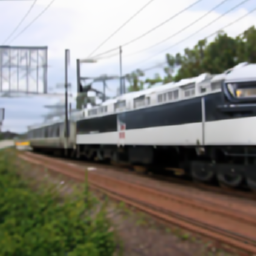}
   & \includegraphics[scale=0.22]{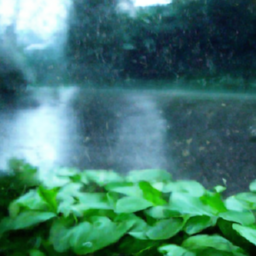} & \includegraphics[scale=0.22]{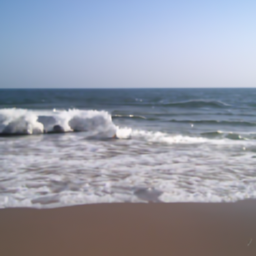} \\
   
    A photo of a cat meowing & A photo of a moving train &  A photo of raindrops falling & A photo of waves on the beach \\ 
\end{tabular}
\caption{Results of audio-to-image generation. The top row was generated from audio clips (accessible from the audio icons), and the bottom row was generated by describing the audio clips in text.}
\label{fig:audio-gen}
\end{figure}

We consider the recently proposed task \cite{girdhar2023imagebind} of generating images given audio prompts. The aim is to repurpose an existing text-to-image generative model to be conditioned on audio in lieu of text. \citet{girdhar2023imagebind} achieved this using a private reimplementation of DALLE-2 \cite{ramesh2022hierarchical}.
We opt to use FuseMix to perform this task while only using publicly available models: we use GLIDE\footnote{Stable Diffusion \cite{rombach2022high} was not considered since its text conditioning is high-dimensional making alignment more challenging than with GLIDE.} \cite{nichol2021glide}, a text-conditioned diffusion model which leverages CLIP\footnote{We use GLIDE's reimplementation of CLIP trained on noisy images.} \cite{radford2021clip} to condition on text. We apply our method to align the latent space of Whisper into the latent space of CLIP to endow GLIDE with audio-conditioning capabilities (see details in Appendix \ref{sec:appendix-impl}). In \autoref{fig:audio-gen}, we provide examples of generated samples using various sounds. While we omit quantitative analysis for this task due to a lack of suitable metrics, we provide a qualitative comparison of each sample with a corresponding sample generated from the original text-conditioned GLIDE using a text prompt that is semantically equivalent to the audio prompt. For example, for the sound of a cat meowing, we compare with the text prompt ``a photo of a cat meowing''. We find it noteworthy that conditioning GLIDE on audio prompts using FuseMix can produce samples of similar quality and fidelity as conditioning on text prompts, even though GLIDE itself was never trained with audio data.

\subsection{Ablations}
\begin{figure}[htbp]
  \footnotesize
  \centering  
  \begin{subfigure}[]{0.32\columnwidth}
    \centering
    \hspace{-0.4cm}
    \begin{tikzpicture}
      \begin{axis}[
        ybar,
        width=3.5cm,
        height=3.5cm,
        bar width=0.2cm,
        ylabel={R@1},
        xtick align=inside,
        xlabel style={yshift=0.2cm},
        ylabel style={yshift=-0.7cm},
        xtick=data,
        ymin=40,
        enlarge x limits=0.25,
        nodes near coords={}, 
        nodes near coords align={vertical},
        xticklabels={
          {S},
          {B},
          {L}
        },
        grid=major,
        ]
        \addplot coordinates {(1,54.04) (2,61.96) (3, 71.2)};
      \end{axis}
    \end{tikzpicture}
    \caption{Model Size}
    \label{fig:model-size}
  \end{subfigure}%
  \begin{subfigure}[]{0.33\columnwidth}
    \centering
    \hspace{-0.3cm}
    \begin{tikzpicture}
      \begin{axis}[
        width=3.5cm,
        height=3.5cm,
        ylabel={R@1},
        xlabel style={yshift=0.2cm},
        ylabel style={yshift=-0.7cm},
        xtick={100,1000,2000},
        xticklabels={1k, 10k, 20k},
        grid=major,
        ]
        \addplot[blue, mark=*, mark size=1] coordinates {(64,66.02) (128,67.2) (256,68.34) (512,68.94) (1024,70.1) (2048,71.2)};
      \end{axis}
    \end{tikzpicture}
    \caption{Batch Size}
    \label{fig:batch-size}
  \end{subfigure}%
    \begin{subfigure}[]{0.35\columnwidth}
        \centering
        \setlength{\tabcolsep}{2pt}
        \hspace{1cm}
        \begin{tabular}{ccccccc}
            \hline
            Aug. &  \multicolumn{2}{c}{R@1} \\
            & t $\rightarrow$ i & i $\rightarrow$ t\\
            \hline
            None & 66.1 & 80.5 \\
            \hline
            GN & 66.7 &  81.5 \\
            \hline
            RQ \cite{wu2023randomizedquantization} & 67.8 &  82.6\\
            \hline
            FuseMix & \textbf{71.2} & \textbf{84.8}\\
            \hline
        \end{tabular}
        \captionsetup{margin={0cm,-0.2cm}}
        \caption{Type of Aug.}
        \label{tab:data-aug-method}
    \end{subfigure}
  \label{fig:ablations}
  \caption{Measuring the effect of model size, batch size, and data augmentations on downstream performance, evaluated with the Flickr30k test set. GN denotes Gaussian noise with a standard deviation of 0.01 and RQ denotes random quantization. By default, R@1 denotes text-to-image Recall@1.}
  \vspace{5pt}
\end{figure}
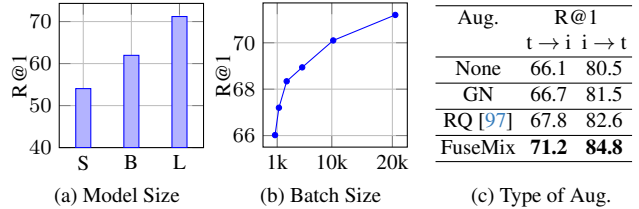

\textbf{Effect of Unimodal Encoder Size.} Given the plug-and-play nature of our method, we would hope that larger underlying unimodal encoders would be beneficial for multimodal fusion. We study this effect by evaluating downstream performance for various sizes of encoders. We consider the following combinations: DINOv2 ViT-S/14 \& BGE Small; DINOv2 ViT-B/14 \& BGE Base; and DINOv2 ViT-G/14 \& BGE Large, referred to as S, B, and L, respectively, in \autoref{fig:model-size}. As shown, scaling the unimodal encoders translates to improved downstream performance.

\textbf{Effect of Batch Size.} As mentioned in Sec. \ref{sec:impl}, since training our fusion adapters requires minimal compute, we can use larger batch sizes even on a single GPU. In \autoref{fig:batch-size}, we see that our method can benefit from more negative samples in the contrastive objective, which is consistent with findings in previous work \cite{tian2020contrastive, he2020momentum, chen2020simclr}.

\textbf{Effect of Data Augmentations.} In \autoref{tab:data-aug-method}, we evaluate the importance of data augmentations and compare our proposed FuseMix with other modality-agnostic data augmentation schemes, namely Gaussian noise and random quantization \cite{wu2023randomizedquantization}. We note that data augmentations generally seem beneficial in our setting, although FuseMix provides the largest improvement in performance, further validating our proposed approach.

%% file: sections/7_conclusion.tex
\section{Conclusion and Future Work}
\label{sec:conclusion]}
In this work, we have proposed a framework for multimodal fusion that is both compute-efficient and data-efficient which can effectively bootstrap from arbitrary pre-trained unimodal encoders. We have introduced FuseMix, a simple yet effective multimodal augmentation scheme on latent space inspired by mixup.  However, while our method can benefit from powerful unimodal encoders, we are limited by the semantic information that they have previously learned \cite{kossen2023three}. It would thus be an interesting future direction to apply efficient fine-tuning methods \cite{hu2021lora, dettmers2023qlora} to the unimodal encoders during fusion, although this would incur additional overhead. We also highlight that since our framework essentially considers unimodal encoders as black box models (i.e. we only use their latent encodings from their penultimate layer), this opens up exciting applications whereby we can perform multimodal fusion with encoders only accessible via an API. 

%% file: sections/appendix.tex
\clearpage
\appendix
\appendixpage

\section{Architecture}
\label{sec:appendix-arch}
For our fusion adapters $h_X$ and $h_Y$, we use a simple inverted bottleneck MLP architecture. To illustrate the simplicity of our design, we provide its pseudocode in Algorithm \ref{algo:mlp}. By default, we use an expansion factor of 4, dropout of 0.6, and a shared latent space of dimension 512. We specify the fusion adapter depths we used for each task in Appendix \ref{sec:appendix-impl}.

\makeatletter
\patchcmd{\@algocf@start}
  {-1.5em}
  {2pt}
  {}{}
\makeatother

\setcounter{algocf}{1}
\DecMargin{0em}
\begin{algorithm}[htbp]
\scriptsize
\SetAlCapNameFnt{\footnotesize}
\SetAlCapFnt{\footnotesize}
\caption{PyTorch-style pseudocode of our fusion adapters.}
\label{algo:mlp}
\SetAlgoLined
    \PyComment{D\_x, D\_y: latent dimension of unimodal encoders} \\
    \PyComment{D\_s: latent dimension of shared space} \\
    \PyComment{depth\_x, depth\_y: number of blocks for each adapter} \\
    \PyComment{expansion\_factor: expansion factor hyperparameter} \\
    \PyComment{dropout: dropout hyperparameter} \\
    \BlankLine
    \PyCode{from torch import nn} \\
    \BlankLine
    \PyCode{class Block(nn.Module):} \\
    \Indp
        \PyCode{def \_\_init\_\_(self, dim, expansion\_factor=4, dropout=0.6):} \\
        \Indp
            \PyCode{super().\_\_init\_\_()} \\
            \PyCode{self.fn = nn.Sequential(} \\
            \Indp
                \PyCode{nn.Linear(dim, int(expansion\_factor * dim)),} \\
                \PyCode{nn.GELU(),} \\
                \PyCode{nn.Dropout(dropout),} \\
                \PyCode{nn.Linear(int(expansion\_factor * dim), dim),} \\
            \Indm
            \PyCode{)} \\
            \PyCode{self.ln = nn.LayerNorm(dim)} \\
        \Indm
        \BlankLine
        \BlankLine
        \PyCode{def forward(self, x):} \\
        \Indp
            \PyCode{return x + self.fn(self.ln(x))} \\
        \Indm
    \Indm
    \BlankLine
    \BlankLine
    \BlankLine
    \PyCode{h\_X = nn.Sequential(} \\
    \Indp
        \PyCode{*[Block(D\_x, expansion\_factor, dropout) for \_ in range(depth\_x)],} \\
        \PyCode{nn.LayerNorm(D\_x),} \\
        \PyCode{nn.Linear(D\_x, D\_s),} \\
    \Indm
    \PyCode{)} \\
    \BlankLine
    \PyCode{h\_Y = nn.Sequential(} \\
    \Indp
        \PyCode{*[Block(D\_y, expansion\_factor, dropout) for \_ in range(depth\_y)],} \\
        \PyCode{nn.LayerNorm(D\_y),} \\
        \PyCode{nn.Linear(D\_y, D\_s),} \\
    \Indm
    \PyCode{)} \\
\end{algorithm}
\setlength{\textfloatsep}{10pt}

\section{Implementation Details}
\label{sec:appendix-impl}
For all experiments, we use the AdamW \cite{loshchilov2018fixing} optimizer during training. We perform learning rate warmup by linearly increasing the learning rate from $10^{-6}$ to \texttt{lr} (which we specify for each task below) during the first epoch. We then decay the learning rate using a cosine
schedule \cite{loshchilov2016sgdr}. We also set our FuseMix Beta distribution hyperparameter as $\alpha=1$ so that the interpolation coefficient is sampled as $\lambda \sim \mathcal{B}(1,1)$.\footnote{$\mathcal{B}(\alpha,\alpha)$ is the uniform distribution when $\alpha=1$, concentrates around $0$ and $1$ when $\alpha<1$, and is unimodal when $\alpha>1$.} We note that when mixup is performed on ambient space, it is common to select small $\alpha$ \cite{zhang2018mixup, verma2021dacl, hao2023mixgen}. This ensures that inputs are only slightly  perturbed so that they remain semantically meaningful. Conversely, in FuseMix, we are operating on the latent space of pre-trained unimodal encoders where we find that relatively larger $\alpha$ can improve performance in our experiments, which suggests that larger perturbations on latent space can remain semantically meaningful (see result in Appendix \ref{sec:appendix-albations}). We next describe specific details and hyperparameters for each task we consider:

\textbf{Image-Text Retrieval.} We use a depth of 4 for both fusion adapters (see ablation in Appendix \ref{sec:appendix-albations}) which we train for 500 epochs with a batch size of $B=20$K. We set the learning rate as  \texttt{lr}$=10^{-3}$ and use weight decay of 0.1 during optimization. 

\textbf{Audio-Text Retrieval.} We use a depth of 2 for both fusion adapters, which we train for 50 epochs with a batch size of $B=2$K. We set the learning rate as \texttt{lr}$=10^{-4}$ and use weight decay of 0.5 during optimization. 

\textbf{Audio-to-Image Generation.} Since we align the latent space of Whisper's encoder into the latent space of CLIP, we are treating CLIP's latent space as our shared space. This means that we only require one fusion adapter to map from Whisper space into CLIP space  -- for which we use a depth of 2. We note that this does not require any changes to our framework since it is equivalent to setting one of our fusion adapters as the identity network in Algorithm \ref{algo:mixup}. For this experiment, we use 50K audio-text pairs from the AudioCaps \cite{audiocaps} training set and a 50K subset of AudioSet \cite{gemmeke2017audioset}. Other hyperparameters are identical to those for audio-text retrieval. During inference, we can therefore map audio inputs to CLIP space and treat them as though they were CLIP text latents, which GLIDE can then use for conditioning.

\section{Additional Ablations}
\label{sec:appendix-albations}

\begin{figure}[htbp]
  \footnotesize
  \centering  
  \begin{subfigure}[]{0.45\columnwidth}
    \centering
    \hspace{-1cm}
    \begin{tikzpicture}
      \begin{axis}[
        width=3.5cm,
        height=3.5cm,
        ylabel={R@1},
        xlabel style={yshift=0.2cm},
        ylabel style={yshift=-0.7cm},
        ymin=61,
        ymax=74,
        grid=major,
        ]
        \addplot[blue, mark=*, mark size=1] coordinates {(0,66.1) (0.2,68) (0.4,68.8)(0.6,69.3) (0.8,70.4) (1,71) (1.5, 70.8)  (2,70.6)};
      \end{axis}
    \end{tikzpicture}
    \vspace*{-0.1cm}
    \caption{$\alpha$}
    \vspace*{0.2cm}
    \label{fig:alpha}
  \end{subfigure}
  \begin{subfigure}[]{0.45\columnwidth}
    \centering
    \hspace{-0.9cm}
    \begin{tikzpicture}
      \begin{axis}[
        width=3.5cm,
        height=3.5cm,
        ylabel={R@1},
        xlabel style={yshift=0.2cm},
        ylabel style={yshift=-0.7cm},
        ymin=61,
        ymax=74,
        grid=major,
        ]
        \addplot[red, mark=*, mark size=1] coordinates {(1, 64.88) (2, 68.46) (3, 69.58) (4, 71.2) (5, 70.5)};
      \end{axis}
    \end{tikzpicture}
    \vspace*{-0.1cm}
    \caption{Adapter Depth}
    \label{fig:num-layers}
  \end{subfigure}
  \label{fig:appendix-ablate}
  \caption{Text-to-image results evaluated on the Flickr30k test set.}
\end{figure}
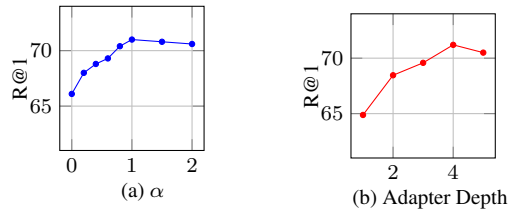 

We provide results for a few additional ablations. First, we observe in \autoref{fig:alpha} that our method can generally benefit from larger $\alpha$ (see Appendix \ref{sec:appendix-impl} for a relevant discussion). We also find in \autoref{fig:num-layers} that as the fusion adapters deepen, performance gradually increases until a depth of 4 where performance peaks. These results validate our setting of these hyperparameters detailed in Appendix \ref{sec:appendix-impl}.

\section{Determinantal Point Processes}
\label{sec:appendix-dpp}

We begin with a brief summary of determinantal point processes (DPPs) for completeness, and refer readers to \cite{kulesza2012determinantal} for a thorough overview of DPPs in machine learning. Consider the set $\mathcal{I} \triangleq \{1, 2, \dots, N\}$, which should be understood as the set of indices of a dataset $\{z_i\}_{i \in \mathcal{I}} \subset \mathcal{Z}$ with $N$ distinct elements. Consider also a symmetric positive semi-definite $N \times N$ matrix $L$, such that $L_{ij}$ measures similarity between $z_i$ and $z_j$. A common choice for this matrix is to specify a positive semi-definite kernel $K: \mathcal{Z} \times \mathcal{Z} \rightarrow \mathbb{R}$ and set $L_{ij} = K(z_i, z_j)$.\footnote{Recall that $K$ is a positive semi-definite kernel if, for every $N$ and every finite subset $\{z_i\}_{i \in \mathcal{I}}$ of $\mathcal{Z}$ of size $N$, the corresponding $N \times N$ matrix $L$ is always positive semi-definite.} A DPP is a distribution over subsets of $\mathcal{I}$, where the probability of obtaining $S \subset \mathcal{I}$ is given by
\begin{equation}
    p_S\left(S \right) = \dfrac{\det L_S}{\displaystyle \sum_{S' \subset \mathcal{I}} \det L_{S'}},
\end{equation}
where $L_S$ corresponds to the $|S| \times |S|$ submatrix of $L$ whose row and column indices are given by $S$. The idea behind DPPs is that diverse subsets are more likely to be sampled, where diversity is measured through dissimilarity (as specified in $L$) of the elements in $\{z_i\}_{i \in S}$. DPPs can be extended to $k$-DPPs \cite{kulesza2011kdpp}, where an integer $k$ is specified and the constraint is added that $S$ must have exactly $k$ elements, or more formally
\begin{equation}
    p_S\left(S \mid |S|=k\right) = \dfrac{\det L_S}{\displaystyle \sum_{\substack{S' \subset \mathcal{I} \\ |S'| = k}} \det L_{S'}}\mathds{1}\left(|S| = k\right),
\end{equation}
where $\mathds{1}(\cdot)$ denotes an indicator function. In the DPP literature, it can be of interest to find a mode of a DPP or $k$-DPP (i.e.\ finding ``maximally diverse'' subsets, potentially of specified size $k$) rather than to sample from these distributions. In our case, we follow the greedy algorithm proposed in \cite{chen2018fast}, whose goal is to obtain a mode $S^*$ of a $k$-DPP:
\begin{equation}\label{eq:dpp_obj}
    S^* \in \argmax_{\substack{S \subset \mathcal{I}\\ |S| = k}} \quad  \det L_S.
\end{equation}
To specify $L$, we first considered the kernel $K(z, z') = z \cdot z'$ in an attempt to leverage the prior knowledge that cosine similarity is sensible on the latent space $\mathcal{Z}$ of pre-trained encoders.\footnote{In our experiments, we subsampled 75K (i.e. $N=75$K) image-text pairs from the COCO dataset to ensure $L$ was able to fit in memory, and took $\mathcal{Z}$ as the latent space of the BGE text encoder.} However, the resulting matrix $L$ has low rank -- at most the dimension of $\mathcal{Z}$ -- and a requirement for the $\argmax$ in \autoref{eq:dpp_obj} to not be the empty set is that $k \leq \rank(L)$. To be able to use larger $k$, we thus changed the kernel to $K(z, z') = (z \cdot z' + 1)^2$, which is monotonically increasing in $z \cdot z'$, but results in an $L$ with much larger rank. We emphasize that in our work we are using $k$-DPPs only to evaluate the effect of dataset diversity for various values of $k$ (i.e. various subset sizes) rather than suggesting its use to curate diverse datasets in practice, which would be too costly.